\documentclass[letterpaper, 10 pt, conference]{ieeeconf}  

\IEEEoverridecommandlockouts                              

\renewcommand{\baselinestretch}{0.95}

\title{\LARGE \bf
Stretchable Capacitive and Resistive Strain Sensors: \\Accessible Manufacturing Using Direct Ink Writing}

\author{Lukas Cha$^{1,2}$, Sonja Groß$^{3,4}$, Shuai Mao$^{1}$, Tim Braun$^{3}$, Sami Haddadin$^{3,4}$, and Liang He$^{1,2}$
\thanks{*This work was supported by The Podium Institute for Sports Medicine and Technology, University of Oxford and the German Research Foundation (DFG, Deutsche Forschungsgemeinschaft) as part of Germany’s Excellence Strategy – EXC
2050/1 – Project ID 390696704 – Cluster of Excellence
“Centre for Tactile Internet with Human-in-the-Loop” (CeTI) of Technische Universität Dresden.} 
\thanks{$^{1}$Institute of Biomedical Engineering,
        University of Oxford, United Kingdom
        {\tt\small}}%
\thanks{$^{2}$The Podium Institute for Sports Medicine and Technology,
        University of Oxford, United Kingdom
        {\tt\small}}%
\thanks{$^{3}$Munich Institute of Robotics and Machine Intelligence,
        Technical University of Munich, Germany
        {\tt\small}}%
\thanks{$^{4}$The Centre for Tactile Internet with Human-in-the-Loop (CeTI), Dresden University of Technology, Germany
        {\tt\small}}%
\thanks{All inquiries can be addressed to: \texttt{lukas.cha@eng.ox.ac.uk}
        {\tt\small}}%
}

\usepackage{graphicx}
\usepackage{amsmath}
\usepackage{xcolor}

\begin{document}

\maketitle
\thispagestyle{empty}
\pagestyle{empty}

\begin{abstract}
As robotics advances toward integrating soft structures, anthropomorphic shapes, and complex tasks, soft, highly stretchable mechanotransducers are becoming essential. To reliably measure tactile and proprioceptive data while ensuring shape conformability, stretchability, and adaptability, researchers have explored diverse transduction principles alongside scalable and versatile manufacturing techniques. Nonetheless, many current methods for stretchable sensors are designed to produce a single sensor configuration, thereby limiting design flexibility. Here, we present an accessible, flexible, printing-based fabrication approach for customisable, stretchable sensors. Our method employs a custom-built printhead integrated with a commercial 3D printer to enable direct ink writing (DIW) of conductive ink onto cured silicone substrates. A layer-wise fabrication process, facilitated by stackable trays, allows for the deposition of multiple liquid conductive ink layers within a silicone matrix. To demonstrate the method’s capacity for high design flexibility, we fabricate and evaluate both capacitive and resistive strain sensor morphologies. Experimental characterisation showed that the capacitive strain sensor possesses high linearity ($R^2 = 0.99$), high sensitivity near the 1.0 theoretical limit ($GF = 0.95$), minimal hysteresis ($DH = 1.36\%$), and large stretchability (550\%), comparable to state-of-the-art stretchable strain sensors reported in the literature.

\end{abstract}

\vspace{-3mm}
\section{INTRODUCTION}

As robots increasingly enter unstructured environments, soft and stretchable mechanotransducers become increasingly important \cite{haddadin2018tactile} for tactile sensing and proprioception in applications such as soft robotic manipulators \cite{he2020soft}, wearables \cite{sundaram2019}, prosthetics \cite{Donati2024}, and human-robot interaction \cite{olugbade2023touch}. These sensors enable the detection of contact points, force, and shear, as well as derivative measurements like object shape, joint angles and object slippage during grasping \cite{ Wang2023}. Skin-like sensors must exhibit high stretchability, shape conformity, scalability, multi-modal sensitivity, and repeatability \cite{tan2020soft,Dahiya2019a}. Current approaches typically achieve stretchability by integrating inherently elastic sensor materials or utilising elastomeric substrates combined with conductive components like liquid metals or composites \cite{Truby2018}. Recent advances include sensors with a stretchability up to 800\% \cite{Hua2018}, various transduction mechanisms \cite{Qu2023} — such as resistive \cite{tan2020soft}, capacitive \cite{Gruebele2020}, optical \cite{Taylor2021}, and pneumatic \cite{he20223d} methods — and the miniaturisation and automation of manufacturing processes through 3D printing technologies \cite{SenthilKumar2019, Gross2024}. For example, Zhang et al. \cite{zhang2023dual} introduced a novel printing approach using direct ink writing of conductive ink onto a rotating substrate to create parallel helical electrodes for capacitive strain sensors. However, this setup exhibited limited design flexibility beyond the helical sensor structure. A highly flexible method called ``Embedded 3D Printing" for resistive sensors was presented by Muth et al. \cite{muth2014embedded} and later adapted to various object shapes by Groß et al. \cite{GrossandHidalgo2023}. On one hand, this method enabled extruding conductive ink into uncured silicone, allowing for highly customisable sensor morphologies. On the other hand, it required intricate tuning of material properties and printing parameters, which limited its ease of use and accessibility. Similarly, the printing-based fabrication method by Ma et al. \cite{ma2019highly}, which utilised printed silicone with embedded graphene sheets, lacked the capability to create multiple conductive layers separated by dielectric layers, thereby limiting its design flexibility for complex sensor architectures. Hence, many existing systems still face limitations in combining high stretchability with accessibility and adaptability to specific applications due to expensive materials such as eutectic gallium indium \cite{chen2020superelastic}, complex system design or manufacturing \cite{Hua2018, Sun2022, Dahiya2019a, Uskin}. 

\begin{figure}[!t]
    \centering
    \includegraphics[width=0.95\columnwidth]{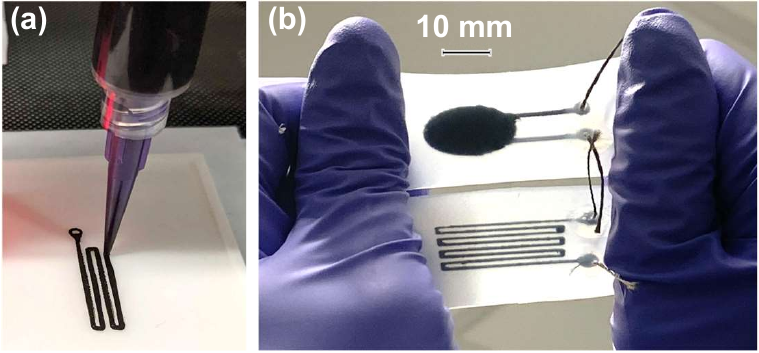}
    \caption{\textbf{(a)} Printing of the conductive pattern. \textbf{(b)} The stretchable capacitive (top) and resistive (bottom) strain sensors.}
    \label{fig:title_figure}
    \vspace{-1.5\baselineskip}
\end{figure}

We therefore propose a manufacturing approach (see Fig. \ref{fig:title_figure}(a)) utilising a commercial 3D printer with an accessible custom-built printhead to manufacture highly stretchable skin-like sensors based on silicone and liquid conductive ink (see Fig. \ref{fig:title_figure}(b)). Our contribution is threefold: (1) we introduce an accessible method for manufacturing conductive ink and silicone-based sensors; (2) we propose sensor designs targeting strain sensing, leveraging resistive and capacitive transduction methods; and (3) we conduct comparative evaluations of these transduction methods through static strain experiments providing a detailed characterisation of the proposed transducers. 
This work aims to offer researchers an accessible platform for direct ink writing (DIW) of stretchable resistive and capacitive strain sensors, promoting ease of use and customisation, and thereby lowering barriers for future advancements in stretchable sensor development for applications such as soft robotics, wearables, and prosthetics.

\section{Materials and Methods}
We designed sensors for two transduction methods and manufactured them utilising a DIW custom-built printer. The following section describes the sensor design, the custom-built printer setup, the overall manufacturing process, and experimental setup.
\subsection{Sensor Design}
\label{section:sensors}
Both sensor principles utilised a two-material concept consisting of a stretchable silicone substrate and liquid conductive ink for the functional layers. 
We utilised Dragonskin-10 FAST (Smooth-On, USA) as a substrate due to its high stretchability up to 1000\% and minimal curing time that still allows for degassing. For the functional layers, we chose carbon conductive grease (846-1P, MG Chemicals, Canada) as it offers accessibility, good printability due to its shear thinning behaviour and chemical compatibility with silicone \cite{muth2014embedded}. Additionally, carbon grease remains liquid within the silicone layers, enabling high stretchability of the developed sensor.

The capacitive strain sensor operated based on a parallel-plate configuration as previously proposed by \cite{hu2022super}. Applied strain leads to a change in distance $d$ of the two conductive layers and therefore a change in capacitance $C$ via the relation 
\vspace{-1.5mm}\cite{souri2020wearable}: \begin{equation}
    \label{eq:capacitance}
    C = \frac{\varepsilon_0 \varepsilon_r A}{d}
    \end{equation}
    
where $\varepsilon_0$ is the permittivity of free space, $\varepsilon_r$ is the relative permittivity of the dielectric material, $A$ is the area of the circular electrodes with 12 mm diameter in the unstrained state.

\begin{figure}[!ht]
    \centering
    \includegraphics[width=0.6\columnwidth]{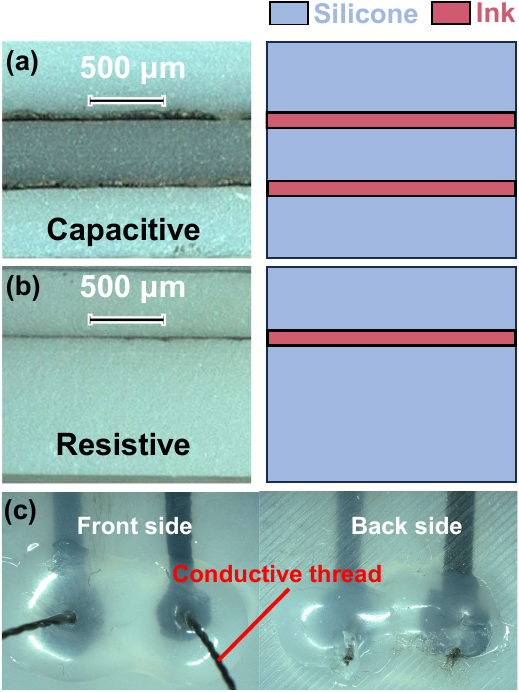}
    \caption{\textbf{(a)} Capacitive strain sensor cross-section with two conductive ink layers shown in photo and schematic. \textbf{(b)} Resistive strain sensor cross-section with a single conductive ink layer in photo and schematic.\textbf{(c)} Front and back side of a capacitive strain sensor, showing the electrical interface utilising conductive thread.}
    \label{fig:cross_section}
    \vspace{-1.0\baselineskip}
\end{figure}

For resistive strain sensing, a single-layer printed strain gauge pattern, decreasing the signal-to-noise-ration was used as previously proposed by \cite{muth2014embedded, GrossandBreimann2022}. This strain gauge pattern had a dimension of 10 mm width, 20 mm length, 0.5 mm line width and 1.4 mm line separation. Deformation of the conductive path in length $L$ or cross-section $A_{r}$ leads to changes in electrical resistance $R$, as determined by \cite{souri2020wearable}:
\vspace{-1.5mm}
\begin{equation}
\label{eq:resistive}
    R = \rho\frac{L}{A_{r}}
\end{equation}

where $\rho$ is the resistivity of the sensing material. Both sensor prototypes exhibited the same thickness of 2.5 mm and were cut to the same dimensions of 25 mm by 60 mm. The thickness of 2.5 mm resulted from three silicone layers, featuring two 0.75 mm outer layers and a 0.5 mm inner layer, visible in Fig. \ref{fig:cross_section}(a).

\subsection{Printer System}
\begin{figure}[!t]
    \centering\includegraphics[width=0.9\columnwidth]{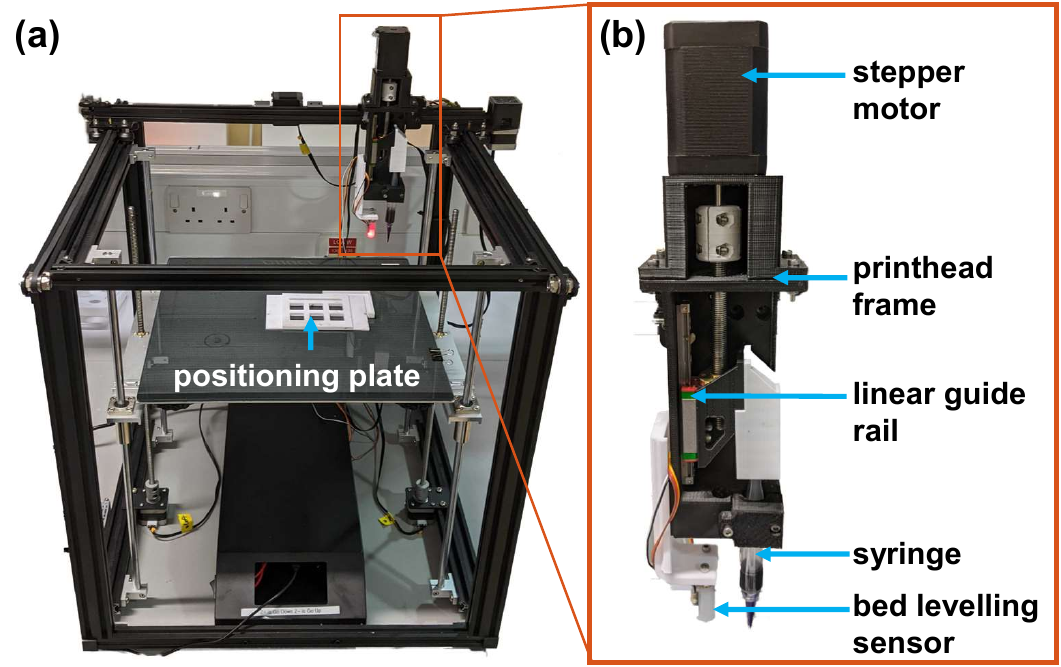}
    \caption{Printer setup. \textbf{(a)} The printer setup with the modified printhead. \textbf{(b)} The modified printhead.}
    \label{Fig_2_printer_setup}
    \vspace{-1.5\baselineskip}
\end{figure}

The printer system (see Fig. \ref{Fig_2_printer_setup}) consisted of a commercial 3D printer (Ender 5 Plus, Creality, China) combined with a custom-built DIW printhead. The printhead was based on the motor-driven syringe pump principle \cite{davila2022open, klar2019ystruder}, as depicted in Fig. \ref{Fig_2_printer_setup}(b), alongside the printer in Fig. \ref{Fig_2_printer_setup}(a). A stepper motor (NEMA 17, Stepperonline, USA) mounted at the printhead frame drove a lead screw, which was connected to the piston of the 5 mL plastic syringe secured at the bottom of the frame. This assembly was secured to a linear guide rail, constraining it to vertical translational movement. Rotation of the lead screw induced a downward translational movement of the assembly, pushing the syringe piston into the syringe barrel. Through this mechanism, the stepper motor pushed the syringe piston, extruding the liquid within it through the nozzle (25-gauge, BENECREAT, China), secured to the syringe via Luer-lock. A 25-gauge nozzle with an inner diameter of 0.515 mm was selected, as it allowed for print lines of the same width. Mounted next to the nozzle head was a bed levelling sensor (BLTouch, South Korea), which calibrated the distance between the nozzle tip and the print bed.

\begin{figure*}[!t]
    \centering
    \includegraphics[width=0.625\textwidth]{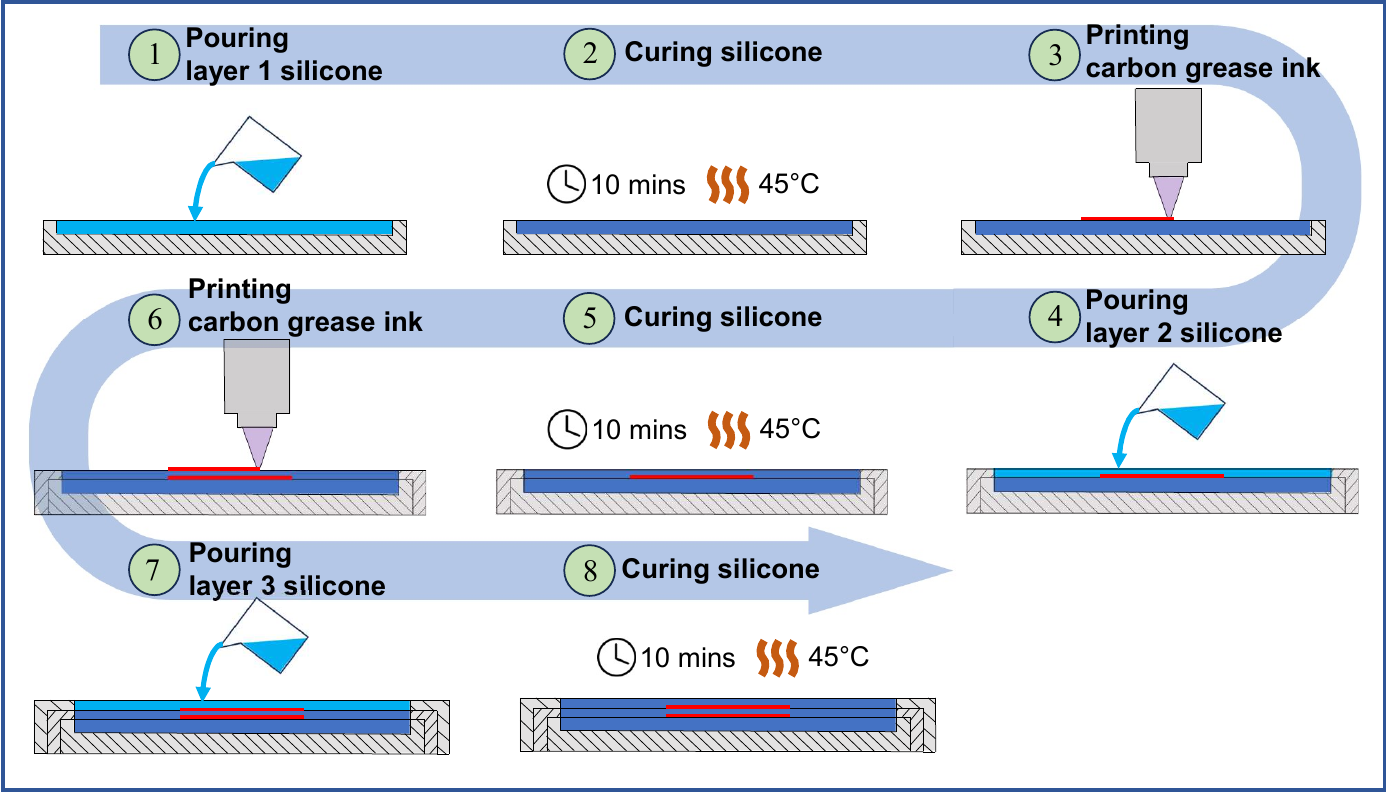}
    \caption{Additive manufacturing fabrication process for capacitive and resistive sensors, consisting of three main steps: pouring silicone, curing silicone, and printing conductive ink. The process used a stackable tray system to facilitate layer-wise assembly, allowing for control over silicone layer thickness and the integration of conductive ink. For capacitive sensors, a dielectric silicone layer was added between conductive layers, while resistive sensors required a single conductive layer encapsulated in silicone.}
    \label{fig:fabrication_process}
    \vspace{-1.5\baselineskip}
\end{figure*}
To control the printer, a Raspberry Pi 3B (Raspberry Pi Foundation, UK) running the open-source Klipper 0.12.0 operating system was used. This operating system provided extensive control over printing parameters and enabled flexible uploading of custom machine code (G-code) from a PC, which communicated with the Raspberry Pi over a wireless network. The G-code was obtained by slicing (Cura 5.4.0, Ultimaker, The Netherlands) the 3D CAD model of the desired pattern with the key parameters: 0.515 mm line width, 100 \% infill in a grid pattern, 0 wall line count, 5 mm/s print speed and retraction enabled. This capability allows for the design of versatile sensor patterns, showcasing the design freedom achievable with our printing-based method. While the Raspberry Pi handled high-level control, low-level control of the stepper motors was performed by the printer mainboard, and the two were connected via USB.


\subsection{Fabrication Process}
The fabrication procedure as depicted in Fig. \ref{fig:fabrication_process} was composed of four main steps (steps 1-4) that were repeated depending on the number of conductive ink prints and the corresponding silicone layers desired (steps 5-8). Here, all the steps are explicitly shown for the fabrication of a capacitive strain sensor with our method.
We utilised a tray system that enabled layer-wise fabrication. These stackable, custom-designed polyactic acid (PLA) trays (printed with a commercial 3D printer (Prusa XL, Prusa, Czech Republic)) were designed so that the base tray functions as the casting mould for the first silicone layer and the subsequent trays extend the edges of the base tray to allow silicone layers of controlled thickness to be added. 
In step one, we prepared equal parts of the A and B components of the silicone with 10 wt\% of a thinning agent (Thinner, Smooth-On, USA)  for decreased viscosity (preventing smearing of the carbon grease during the fabrication process), mixed and then degassed them utilising a vacuum pump (BA-1, BACOENG, USA). The mixture was then poured to fill the base tray of depth 0.75 mm while any excess silicone was manually scraped off. In step two, the silicone cured for approximately 10 minutes at $45~^\circ\mathrm{C}$ on a hot plate (Ender 3, Creality, China). In step three, the tray with the cured silicone was secured in position on the print bed of the printer via a positioning plate shown in Fig. \ref{Fig_2_printer_setup}(a) and a thin 0.15 mm layer of conductive carbon grease was deposited onto the silicone with the desired pattern. To encapsulate the printed conductive ink, the second tray was placed onto the base tray and uncured silicone was poured, fully covering the ink and adding a second layer of silicone with a thickness defined by the dimensions of the tray. For capacitive sensors, this second silicone layer was the dielectric layer and its thickness (chosen here to be 0.5 mm) determined the base capacitance of the sensor. To complete the capacitive sensor, the previous steps two and three were repeated and the silicone was cured, upon which the second conductive ink layer was deposited and sealed in the final two steps, where the third silicone layer of 0.75 mm thickness was added. For resistive strain  sensors, the sensor was already complete at stage five, as only a single print of the conductive ink was required. To maintain comparable mechanical properties, however, our resistive strain sensor was also fabricated with three silicone layers to reach the overall sensor thickness of 2.5 mm.

Electrical interfacing of the sensor was enabled via conductive thread (Adafruit, USA), which allows the soft character of the overall sensor to be maintained. These were threaded through with a needle at the position of the pads in the printed pattern and secured with a few drops of Dragonskin-10 FAST on both sides, which also prevented the ink from leaking out. The mechanical connection of the conductive thread on both sides of the sensor is shown in Fig. \ref{fig:cross_section}(c).
\vspace{-2mm}
\subsection{Experiments}
\label{section:sensor_char_proc}
\subsubsection{Experimental Procedure}
To characterise the strain sensing performance and stretchability of the capacitive and resistive sensors, two experiments were performed and we computed the sensors' hysteresis, linearity, sensitivity and stretchability. First, we conducted a cyclic strain experiment involving straining the sensor to 300\% of its length and recording the measured capacitance and resistance values. The experiment included five cycles of stretching and releasing.  In addition to the cyclic strain experiment, we conducted a strain-to-failure experiment, straining the sensors until mechanical failure. 



\subsubsection{Experimental Setup}
\begin{figure}[!t]
    \centering
    \includegraphics[width=0.8\columnwidth]{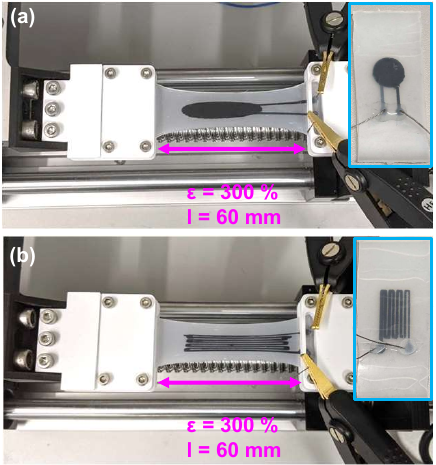}
    \caption{Experimental setup for strain testing. \textbf{(a)} Capacitive strain sensor under strain. The sensor in its unstrained state is shown on the right. \textbf{(b)} Resistive strain sensor under strain. The sensor in its unstrained state is shown on the right.}
    \label{fig:experimental_setup}
    \vspace{-1.5\baselineskip}
\end{figure}

The experimental setup (see Fig. \ref{fig:experimental_setup}) shows the capacitive and resistive strain sensors in their stretched states, fastened at both ends with custom-built 3D-printed clamps (printed with a commercial 3D printer (Prusa XL, Prusa, Czech Republic)). To precisely stretch the sensors to the desired strain level, a motorised guide rail was used, which was controlled by an Arduino Uno (Smart Projects, Italy).

\subsubsection{Data Acquisition and Processing}
To measure and record the capacitance and resistance outputs at different strain, an LCR metre  (LCR 6020, GW-Instek, Taiwan) was used, which was controlled by the PC via USB connection. 

The experiments performed were static strain experiments, where the sensor was first strained to the desired level and kept at that strain level for three seconds as ten capacitance or resistance measurements were recorded by the LCR metre at 10 Hz, triggered by a MATLAB 2022 (Mathworks, USA) script. 

The mean of these ten measurements was used as the data point. To compute the relative change in capacitance $\Delta C/C$ or resistance $\Delta R/R$, the baseline capacitance and resistance in the unstrained state were computed as the mean of 100 measurements. This procedure was repeated at the start of each cycle in the cyclic strain experiment, with curves computed for each cycle using the corresponding baseline measurement.

\subsubsection{Sensor Metric Computation}

To quantify the hysteresis of the sensors, the degree of hysteresis \cite{huang2023high} can be computed:
\vspace{-2mm}
\begin{equation}
    \label{eq:hysteresis_def}
    \text{DH} = \frac{A_{\text{stretch}} - A_{\text{release}}}{A_{\text{stretch}}} \times 100 \%
\end{equation}

where $A_{\text{stretch}}$ is the area under the stretching curve and $A_{\text{release}}$ is the area under the release curve. The degree of hysteresis is computed for each of the five stretch and release cycles and the mean is computed.

To compute the linearity and sensitivity of the strain sensors, a linear curve was first fitted with MATLAB 2022 (Mathworks, USA) to the five-cycle mean relative capacitance or resistance at each strain level, shown in Fig. \ref{fig:linear_fits} with the linear curve and the data points, respectively. From the linear curve, the linearity was obtained as the coefficient of determination $R^2$ and the sensitivity, or gauge factor $GF$, was obtained from the gradient. 

For sensor stretchability, the strain value at the point of mechanical failure was taken.

\section{RESULTS AND DISCUSSION}
\label{section:results & discussion}
We manufactured $7$ capacitive and resistive strain sensors each, and evaluated the fabrication repeatability of our proposed printing setup and the sensor performance in two static strain experiments.

\subsection{Fabrication Repeatability}

Table \ref{table:fabrication_repeatability} shows the zero resistance and zero capacitance recorded for the $7$ fabricated sensors in their unstrained state. The mean and standard deviation of the resistances was $213.14 \pm 59.31$ k$\Omega$, while for the capacitances it was $6.71 \pm 0.42$ pF. The standard deviation of the zero resistance relative to the mean base resistance ($27.8\%$) was larger compared to that of the capacitive sensors ($6.2\%$), indicating a greater variability in the manufacturing of resistive strain sensors compared to the capacitive strain sensors. 

One contributing factor is the precision of our custom printhead. Since electrical resistance directly correlates with the line widths of the printed strain gauge pattern (see Eq. \ref{eq:resistive}), small variations in the extrusion rate of our printhead influence the sensor's zero resistance. This inconsistency in extrusion rate has a smaller effect on capacitive strain sensors, as they use a planar electrode where line width does not directly impact zero capacitance. Future work should focus on optimising the printhead design to ensure a more consistent extrusion rate, thereby enhancing sensor fabrication repeatability.

\vspace{-3mm}
\begin{table}[ht]
\centering
\caption{Zero Values for Resistive and Capacitive Sensors}
\begin{tabular}{|c|c|c|}
\hline
\textbf{Sensor} & \textbf{Resistive (k$\Omega$)} & \textbf{Capacitive (pF)} \\ \hline
1 & 200 & 6.97 \\ \hline
2 & 167 & 6.95 \\ \hline
3 & 140 & 7.00 \\ \hline
4 & 265 & 6.90 \\ \hline
5 & 230 & 6.10 \\ \hline
6 & 180 & 6.10 \\ \hline
7 & 310 & 6.95 \\ \hline
\end{tabular}
\label{table:fabrication_repeatability}
\vspace{-1.0\baselineskip}
\end{table}

\subsection{Strain Sensing Characterisation}

This section presents the results of the cyclic strain experiment and strain-to-failure experiment indicating the sensors' hysteresis, linearity, sensitivity and stretchability characteristics.

\subsubsection{Hysteresis}
\begin{figure}[htb]
    \centering
    \includegraphics[width=0.9\columnwidth]{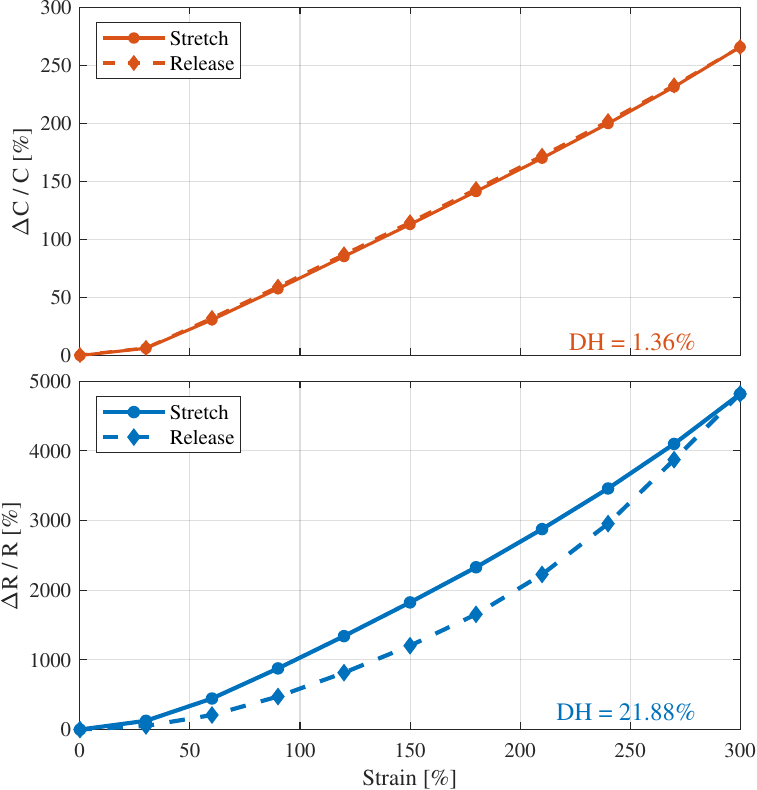}
    \caption{Capacitance and resistance output recorded during the static strain test in the stretch and release phases for the capacitive (top) and resistive strain sensor (bottom), respectively.}
    \label{fig:combined_strain}
    \vspace{-0.5\baselineskip}
\end{figure}

To visualise the hysteresis behaviour of the sensors, the stretch and release curves of one cycle from the cyclic strain test are shown in Fig. \ref{fig:combined_strain}, with the relative capacitance across strain depicted at the top and the relative resistance across strain depicted at the bottom. For the capacitive strain sensor, the stretch and release curves were approximately overlapping, which reflects in a hysteresis of $\text{DH}_{\text{capacitive}} = 1.36 \%$. The resistive strain sensor show a higher hysteresis with $\text{DH}_{\text{resistive}} = 21.88 \%$. 

Our results align with existing literature, showing a comparable hysteresis below 2\% for capacitive strain sensors \cite{souri2020wearable, amjadi2016stretchable} and a higher hysteresis for resistive sensors. Similar levels of hysteresis have been reported for resistive strain sensors incorporating carbon black nanomaterials, such as carbon conductive grease embedded in an elastomeric substrate \cite{rosset2013flexible}, as reflected in our results with $\text{DH}_{\text{resistive}} = 21.88 \%$. This increased hysteresis is attributed to differences in the time scales for the breakdown and reformation processes among carbon particles in the resistive network \cite{rosset2013flexible}, as well as friction and limited interfacial adhesion between the carbon active material and the substrate \cite{amjadi2014highly}. Future research will aim to enhance these material properties to support more linear sensor behaviour.


\subsubsection{Linearity and Sensitivity}
\begin{figure}[htb]
    \centering
    \includegraphics[width=0.9\columnwidth]{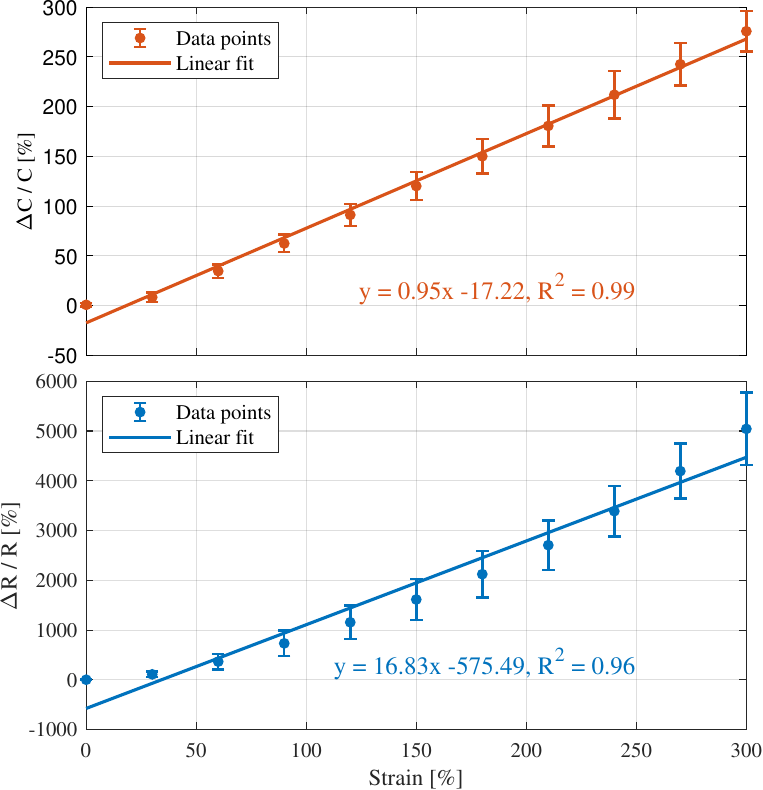}
    \caption{Mean capacitance (top) and resistance (bottom) across the five cycles recorded at various strain levels for the capacitive and resistive strain sensors with the error bars indicating a standard deviation. A linear curve is fitted onto the data points, yielding linearity and sensitivity values.}
    \label{fig:linear_fits}
    \vspace{-1.5\baselineskip}
\end{figure}

Fig. \ref{fig:linear_fits} show the fitted linear curves on the average sensor output over the five cycles, indicating coefficients of determination $\text{R}_{\text{capacitive}}^2 = 0.99$ and $\text{R}_{\text{resistive}}^2 = 0.96$. For the sensitivities, obtained by the gradient, they are $GF_{\text{capacitive}} = 0.95$ for the capacitive strain sensor and $GF_{\text{resistive}} = 16.83$ for the resistive strain sensor.

These findings are in accordance with stretchable strain sensors in literature \cite{souri2020wearable, amjadi2016stretchable}. For the capacitive strain sensor, the $GF$ was close to the theoretical maximum of 1.0, which is limited by the geometrical transduction principle as derived in \cite{yao2015nanomaterial}. The sensitivity of our proposed resistive strain sensor exceeded previously reported sensors utilising carbon grease and silicone with $GF = 3.8$ \cite{muth2014embedded}. The higher linearity exhibited by the capacitive strain sensor compared to the resistive strain sensor also aligns with trends reported in the literature \cite{souri2020wearable, amjadi2016stretchable}.

\subsubsection{Stretchability}

\begin{figure}[h]
    \centering
    \includegraphics[width=0.95\columnwidth]{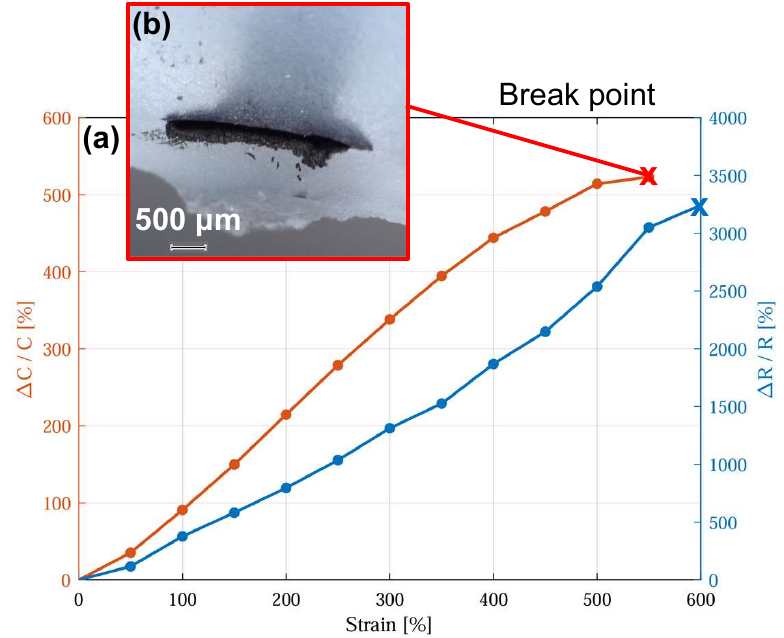}
    \caption{\textbf{(a)} Capacitance and resistance recorded at different strain levels until mechanical failure of the sensor. \textbf{(b)} Break surface of the capacitive strain sensor under microscope.}
    \label{fig:strain_to_failure}
    \vspace{-1\baselineskip}
\end{figure}

Fig. \ref{fig:strain_to_failure}(a) depicts the relative capacitance and resistance outputs of the tested sensors until mechanical failure, with Fig. \ref{fig:strain_to_failure}(b) showing the break surface of the capacitive strain sensor under a microscope. Capacitive and resistive strain sensors experienced mechanical failure at 550\% and 600\% strain, respectively, while sensor linearity was maintained to approximately 400\%. 

This level of stretchability was positioned in the upper mid-range of values reported within the state of the art in the literature \cite{amjadi2016stretchable, souri2020wearable, yao2015nanomaterial}.
\vspace{-2mm}

\section{CONCLUSION}
\label{section:conclusion}

In this study, we implemented and evaluated a low-cost, versatile, printing-based fabrication method for resistive and capacitive, stretchable strain sensors, demonstrating its ease of use and the high design freedom. The fabricated capacitive and resistive strain sensors exhibited favourable characteristics in terms of linearity, sensitivity, and stretchability, with performance metrics comparable to those reported in the literature. This work not only introduces an accessible approach for creating complex sensor configurations, but also highlights the potential of printing-based fabrication in advancing soft, wearable technologies for robotics and health monitoring applications.

Further refinement of the printhead design is needed for increased printing reliability via extrusion rate consistency. Additionally, alternative conductive inks (e.g., silver inks or liquid metals) and flexible substrates compatible with our printer-based method will be explored for enhanced sensor performance as well as tunable sensor characteristics \cite{he2023embodied}. Future work should also validate the scalability of the fabrication method for the printing of high-resolution sensor arrays and extend the printhead to multi-material printing of e.g., the silicone substrate as proposed by \cite{Gross2024}, removing the need to manually pour silicone. By addressing these challenges, this approach could pave the way for highly adaptable, soft sensor technologies suited to diverse fields, including soft robotics, motion tracking, and real-time health diagnostics.
\vspace{-1.5mm}
\section*{Acknowledgments}

The authors would like to thank Massimo Mariello from the Bioelectronics Lab, University of Oxford for his help with the microscope.

\addtolength{\textheight}{-12cm}   








\vspace{-3mm}
\bibliographystyle{IEEEtran} 
\bibliography{references} 

\end{document}